%% file: main.tex
\documentclass[10pt,twocolumn,letterpaper]{article}

\usepackage[pagenumbers]{cvpr} %

\usepackage{graphicx}
\usepackage{amsmath}
\usepackage{amssymb}
\usepackage{booktabs}

\usepackage[table]{xcolor}
\usepackage{multirow}
\usepackage{tabularx}
\usepackage{hhline}
\usepackage{booktabs}
\usepackage{array}
\usepackage[ruled,vlined, noend]{algorithm2e}

\usepackage[utf8]{inputenc}
\usepackage{color}
\usepackage{soul}  %
\usepackage{mdframed} %
\usepackage{siunitx} %

\input{macros}

\renewcommand{\paragraph}[1]{\vspace{1pt}\noindent\textbf{#1.}}
\newcommand{\method}[1]{\textbf{\textit{#1}}}

\usepackage[pagebackref,breaklinks,colorlinks]{hyperref}
\renewcommand*{\backrefalt}[4]{%
    \ifcase #1 \footnotesize{(Not cited.)}%
    \or        \footnotesize{(page~#2)}%
    \else      \footnotesize{(pages~#2)}%
    \fi}

\usepackage[capitalize]{cleveref}
\crefname{section}{Sec.}{Secs.}
\Crefname{section}{Section}{Sections}
\Crefname{table}{Table}{Tables}
\crefname{table}{Tab.}{Tabs.}

\def\cvprPaperID{8087} %
\def\confName{CVPR}
\def\confYear{2023}

\begin{document}

\title{Removing Objects From Neural Radiance Fields}

\author{
Silvan Weder$^{1,2}$\hspace{10pt}
Guillermo Garcia-Hernando$^1$\hspace{10pt}
{\'{A}}ron Monszpart$^1$ \hspace{10pt}
Marc Pollefeys$^2$\hspace{10pt}\\
Gabriel Brostow$^{1,3}$\hspace{10pt}
Michael Firman$^1$\hspace{10pt} 
Sara Vicente$^1$
\\
{\normalsize $^1$Niantic \hspace{60pt} $^2$ETH Zurich \hspace{60pt} $^3$University College London}
\\
{\small \href{https://nianticlabs.github.io/nerf-object-removal}{nianticlabs.github.io/nerf-object-removal}}
}
\maketitle

\definecolor{turquoise}{cmyk}{0.65,0,0.1,0.1} 
\definecolor{purple}{rgb}{0.65,0,0.65}
\definecolor{dark_green}{rgb}{0, 0.5, 0}
\definecolor{orange}{rgb}{0.8, 0.2, 0.2}
\definecolor{red}{rgb}{213, 0, 0}
\definecolor{inpaper}{rgb}{0.5, 0.5, 0.5}
\definecolor{lightblue}{RGB}{209,225,241}

\newcommand{\silvan}[1]{}  
\newcommand{\sara}[1]{} 
\newcommand{\mf}[1]{}
\newcommand{\aron}[1]{}
\newcommand{\ggb}[1]{}
\newcommand{\gb}[1]{}
\newcommand{\todo}[1]{}
\newcommand{\old}[1]{}

\begin{abstract}
Neural Radiance Fields (NeRFs) are emerging as a ubiquitous scene representation that allows for novel view synthesis. Increasingly, NeRFs will be shareable with other people.
Before sharing a NeRF, though, it might be desirable to remove personal information or unsightly objects. Such removal is not easily achieved with the current NeRF editing frameworks.
We propose a framework to remove objects from a NeRF representation created from an RGB-D sequence.
Our NeRF inpainting method leverages recent work in 2D image inpainting and is guided by a user-provided mask.
Our algorithm is underpinned by a confidence based view selection procedure. It chooses which of the individual 2D inpainted images to use in the creation of the NeRF, so that the resulting inpainted NeRF is 3D consistent.
We show that our method for NeRF editing is effective for synthesizing plausible inpaintings in a multi-view coherent manner. We validate our approach using a new and still-challenging dataset for the task of NeRF inpainting.
\vspace{-10pt}
\end{abstract}

\section{Introduction}
\label{sec:intro}

Since the initial publication of Neural Radiance Fields (NeRFs)~\cite{mildenhall2020nerf}, there has been an explosion of extensions to the original framework, \eg \cite{barron2021mipnerf,barron2022mipnerf360,chen2021mvsnerf,kangle2021dsnerf,kim2022infonerf,liu2021editing,Long2022SparseNeuS,mildenhall2020nerf}. %
NeRFs are being used beyond the initial task of novel view synthesis. It is already appealing to get them into the hands of non-expert users for novel applications, \eg for NeRF editing~\cite{yuan2022nerf} or live capture and training~\cite{muller2022instant}. More casual use-cases are driving interesting new technical issues.

\begin{figure}
    \centering
    \includegraphics[width=1.0\columnwidth]{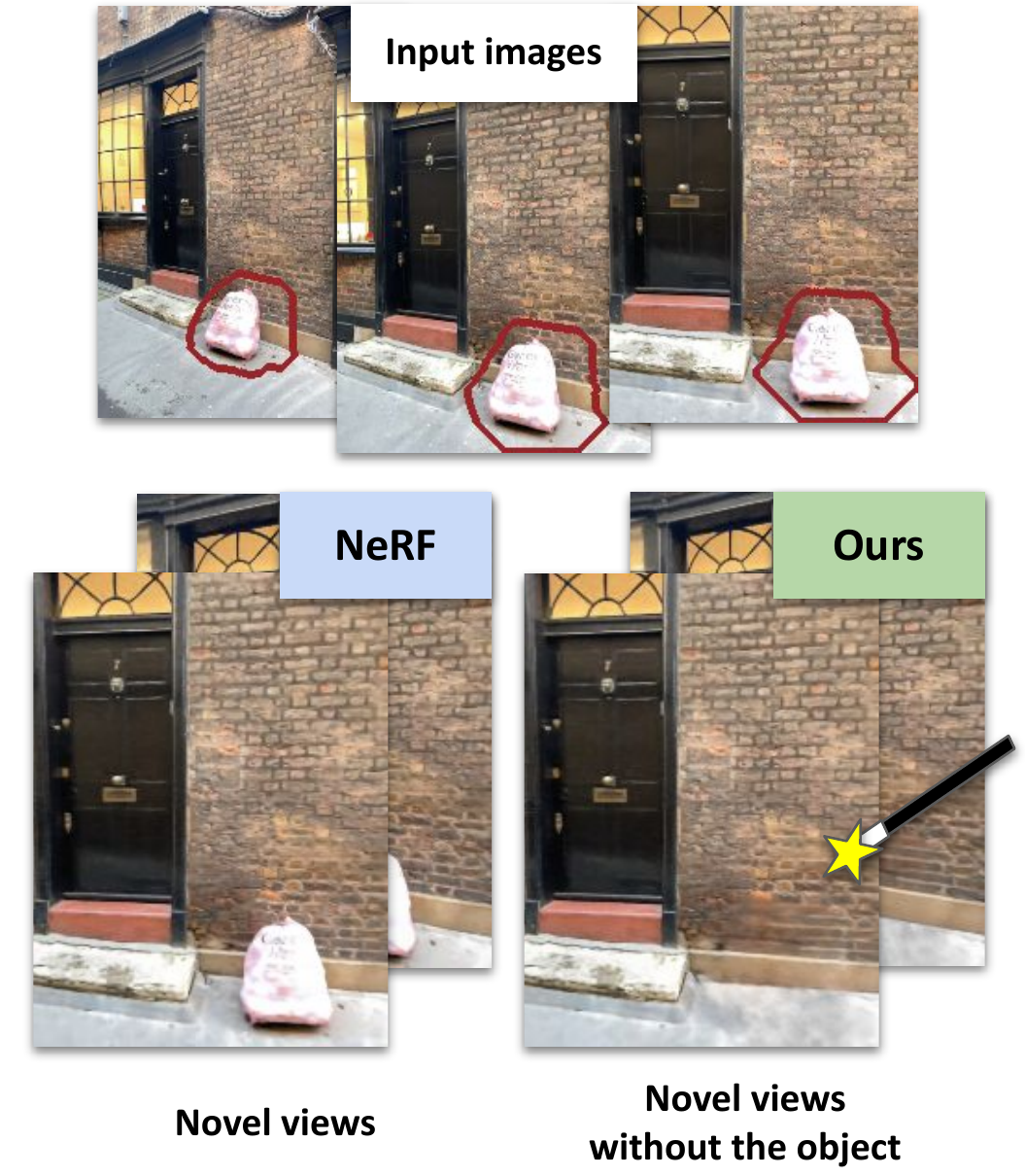}
    \vspace{-0.8cm}
    \caption{\textbf{Removal of unsightly objects.} 
    Our method allows for objects to be plausibly removed from NeRF reconstructions, inpainting missing regions whilst preserving multi-view coherence. \label{fig:teaser}}
    \vspace{-0.4cm}
\end{figure}

One of those issues is how to seamlessly remove parts of the rendered scene. Removing parts of the scene can be desirable for a variety of reasons. For example, a house scan being shared on a property selling website may need unappealing or personally identifiable objects to be removed~\cite{sulaiman2020matterport}. Similarly, objects could be removed so they can be replaced in an augmented reality application, \eg removing a chair from a scan to see how a new chair fits in the environment~\cite{ozturkcan2021service}. Removing objects might also be desirable when a NeRF is part of a traditional computer vision pipeline, \eg removing parked cars from scans that are going to be used for relocalization~\cite{moreau2022lens}.

Some editing of NeRFs has already been explored.
For example, object-centric representations disentangle labeled objects from the background, which allows editing of the trained scene with user-guided transformations~\cite{yang2021objectnerf, wu2022object}, while semantic decomposition allows selective editing and transparency for certain semantic parts of the scene~\cite{kobayashi2022decomposing}.
However, these previous approaches only augment %
information from the input scan, limiting their generative capabilities, \ie the hallucination of elements that haven't been observed from any view.

With this work, we tackle the problem of removing objects from scenes, while realistically filling the resulting holes, as shown in Fig.~\ref{fig:teaser}. 
Solving this problem requires: a) exploiting multi-view information when parts of the scene are observed in some frames but occluded in others and, b) leveraging a generative process to fill areas that are never observed.
To this end, we pair the multi-view consistency of NeRFs with the generative power of 2D inpainting models~\cite{suvorov2022lama} that are trained on large scale 2D image datasets. 
Such 2D inpaintings are not multi-view consistent by construction, and may contain severe artefacts. 
This causes corrupted reconstructions, so we design a new confidence-based view-selection scheme that iteratively removes inconsistent inpaintings from the optimization. 
We validate our approach on a new dataset and show that we outperform existing approaches on standard metrics of image quality, as well as producing multi-view consistent results. 

\textbf{In summary, we make the following contributions:} 1) We propose the first approach focusing on inpainting NeRFs by leveraging the power of single image inpainting. 2) A novel view-selection mechanism that automatically removes inconsistent views from the optimization. 3) We present a new dataset for evaluating object removal and inpainting in indoor and outdoor scenes.

\section{Related work}
\label{sec:relatedw}

\paragraph{Image inpainting}
Image inpainting tackles the problem of plausibly filling in missing regions in a single image.
A typical approach is to use an image-to-image network with an adversarial loss, \eg \cite{pathak2016context,yu2018generative, zeng2019learning,iizuka2017globally}, or with a diffusion model \cite{lugmayr2022repaint}.
Different ways have been proposed to encode the input image, \eg using masked \cite{liu2018image} or Fourier convolutions \cite{suvorov2022lama}.
Image inpainting was extended to also inpaint depth images by \cite{fujii2020rgb}.
However, these methods do not give temporal consistency between video frames, nor the ability to synthesize novel views.

\paragraph{Removing moving objects from videos}
While video inpainting is a well studied problem in computer vision \cite{xu2019VideoInpainting, zeng2020learning}, most works focus on removing moving objects.
This is typically achieved with guidance from nearby frames, \eg via estimating flow \cite{gao2020flow,xu2019VideoInpainting,huang2016temporally}, sometimes using depth \cite{liao2020dvi,bei2020dynamic}.
Perhaps counter-intuitively, moving objects make the task easier, since their movement disoccludes the background, making most parts of the scene visible in at least one of the frames. 
In contrast, to inpaint videos of a static scene a generative approach is required, since large parts of the scene have never been observed. 

\paragraph{Removing static objects from videos}
Where occluded pixels are visible in other frames in the sequence, these can be used to inpaint regions \cite{lepetit2001intuitive,mori2020inpaintfusion, Mori2022InpaintFusion}.
For example, \cite{liu2020learning,liu2021learning} remove static foreground distractors from videos, \eg fences and raindrops.
However, there are typically still pixels which cannot be seen in other views, for which some other method is required to fill them in.
For example, \cite{herling2010advanced} propagates patches from visible pixels into the region to inpaint, and \cite{mori2020inpaintfusion,thonat2016multi} inpaint missing pixels via PatchMatch.
Kim \etal~\cite{kim2021generative} rely on a pre-computed mesh of each scene for object removal.
Our key difference to these methods is that our inpaintings can be extrapolated to novel viewpoints.

\paragraph{Inpainting in novel view synthesis}
Inpainting is often used as a \emph{component} of novel view synthesis, to estimate textures for regions unobserved in the inputs \cite{shih20203d, wang20223d}, \eg for panorama generation \cite{koh2021pathdreamer, hsu2021moving}.
Philip \etal \cite{philip2018plane} enable object removal from image-based-rendering, but with an assumption that background regions are locally planar.

\subsection{Novel view synthesis and NeRFs}

NeRF \cite{mildenhall2020nerf} is a highly popular image-based rendering method which uses a differentiable volume-rendering formulation to represent a scene; a multi-layer perceptron (MLP) is trained to regress the color and opacity given a 3D coordinate and ray viewing direction.
This combined works on implicit 3D scene representations \cite{liu2019learning,park2019deepsdf,mescheder2019occupancy,runz2020frodo,saito2019pifu,saito2020pifuhd,chibane2020implicit}, with light-field rendering \cite{davis2012unstructured} and novel view synthesis \cite{hedman2018deep, liu2019liquid,xu2019view}.
Extensions include work that reduces aliasing artifacts \cite{barron2021mipnerf}, can cope with unbounded scenes \cite{barron2022mipnerf360}, reconstructs a scene from only sparse views \cite{chen2021mvsnerf, Niemeyer2021Regnerf, Long2022SparseNeuS, kim2022infonerf} or makes NeRFs more efficient, \eg by using octrees  \cite{tremblay2022rtmv, yu2021plenoctrees} or other sparse structures \cite{yu2021plenoxels}.

\paragraph{Depth-aware Neural Radiance fields} 
To overcome some of the limitations of NeRFs, particularly the requirement for dense views and the limits in the quality of the reconstructed geometry, depths can be used in training \cite{kangle2021dsnerf, roessle2022dense}. 
These can be sparse depths from structure-from-motion \cite{schonberger2016structure, schoenberger2016mvs}, or depth from sensors \cite{rematas2022urban,sucar2021iMap}.

\paragraph{Object-centric and semantic NeRFs for editing}
One direction of progress in NeRFs is the decomposition of the scene into its constituent objects \cite{Ost_2021_CVPR, yang2021objectnerf, wu2022object}. 
This is done based on motion for dynamic scenes \cite{Ost_2021_CVPR}, or instance segmentation for static scenes \cite{yang2021objectnerf}. 
Both lines of work also model the background of the scene as a separate model. 
However, similar to video inpainting, dynamic scenes allow a better modelling of the background since more of it is visible. 
In contrast, visual artefacts can be seen in the background representation of \cite{yang2021objectnerf, wu2022object}, which model static scenes.
Methods that decompose the scene based on semantics \cite{Zhi2021Inplace, kobayashi2022decomposing} can also be used to remove objects. 
However, they do not try to complete the scene when a semantic part is removed and, for example, \cite{kobayashi2022decomposing} discusses how ``the background behind the deleted objects can be noisy or have a hole because it lacks observation''.

\paragraph{Generative models for novel view synthesis}
3D aware generative models can be used to synthesize views of an object or scene from different viewpoints, in a 3D consistent manner \cite{nguyen2019hologan, schwarz2020graf, devries2021unconstrained, rockwell2021pixelsynth}.
In contrast with NeRF models, which only have a test time component and ``overfit'' to a specific scene, generative models can be used to hallucinate views of novel objects by sampling in the latent variable space.
There has also been some interest in 3D generative models that work for full indoor scenes \cite{devries2021unconstrained, rockwell2021pixelsynth, li2022compnvs}.
However, their capacity to fit the source views (or memorization) can be limited, as shown in the qualitative results of \cite{devries2021unconstrained}.
To train the generative model, \cite{devries2021unconstrained, li2022compnvs, rockwell2021pixelsynth} require a large dataset of indoor scenes with RGB and camera poses and in some cases depth \cite{devries2021unconstrained, li2022compnvs}. In contrast, our use of a 2D pretrained inpainting network, that can be trained on any image, is less dependent on the existence of training data and less constrained to indoor scenarios.

\begin{figure}
    \centering
    \footnotesize
    \centering
    \setlength{\tabcolsep}{1.5pt}
    \begin{tabular}{cc|cc}
        \multicolumn{2}{c|}{Bad Inpaintings} & \multicolumn{2}{c}{Good Inpaintings} \\
        \includegraphics[width=0.24\columnwidth]{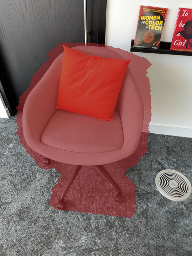} &
        \includegraphics[width=0.24\columnwidth]{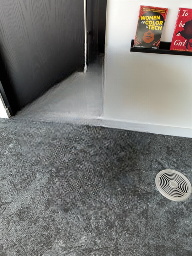} &
        \includegraphics[width=0.24\columnwidth]{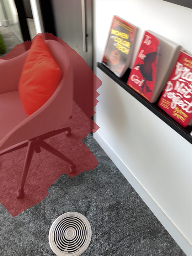} &
        \includegraphics[width=0.24\columnwidth]{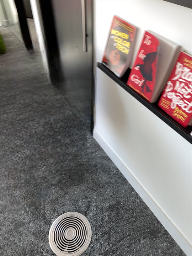} \\
        \includegraphics[width=0.24\columnwidth]{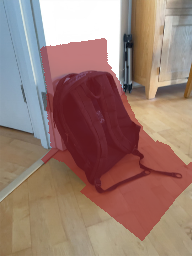} &
        \includegraphics[width=0.24\columnwidth]{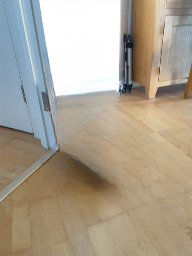} &
        \includegraphics[width=0.24\columnwidth]{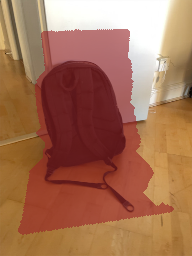} &
        \includegraphics[width=0.24\columnwidth]{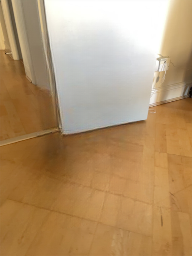}
    \end{tabular}
    \vspace{-5pt}
    \caption{\textbf{Per-frame inpainting} can give plausible results for each frame, but they are not consistent between viewpoints and sometimes contain severe artefacts corrupting the optimization.}
    \label{fig:static-inpainting-inconsistent}
    \vspace{-8pt}
\end{figure}

\begin{figure*}
    \centering
    \includegraphics[width=1.0\textwidth]{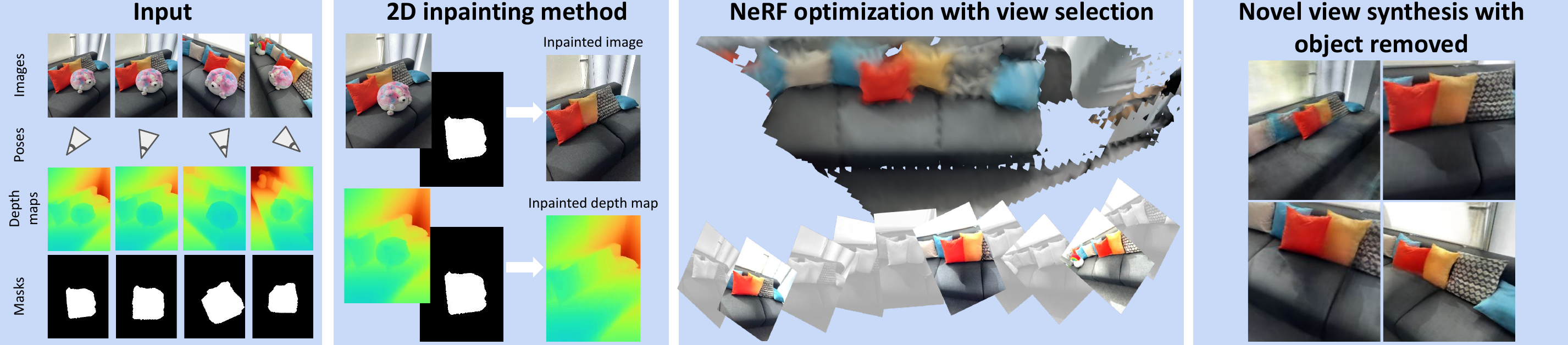}\\
    \caption{\textbf{An overview of our method} We take a sequence of posed RGB-D images together with corresponding 2D masks as input. The 2D frames are inpainted using~\cite{suvorov2022lama} and then used to optimize a neural radiance field. During optimization, our confidence-based view selection automatically removes inconsistent views from the optimization preventing unwanted artefacts in the final result. Finally, novel views can be rendered from the scene, where the object has been removed.}
    \label{fig:method_overview}
    \vspace{-10pt}
\end{figure*}

\begin{figure}
    \centering
    \includegraphics[width=1.0\columnwidth]{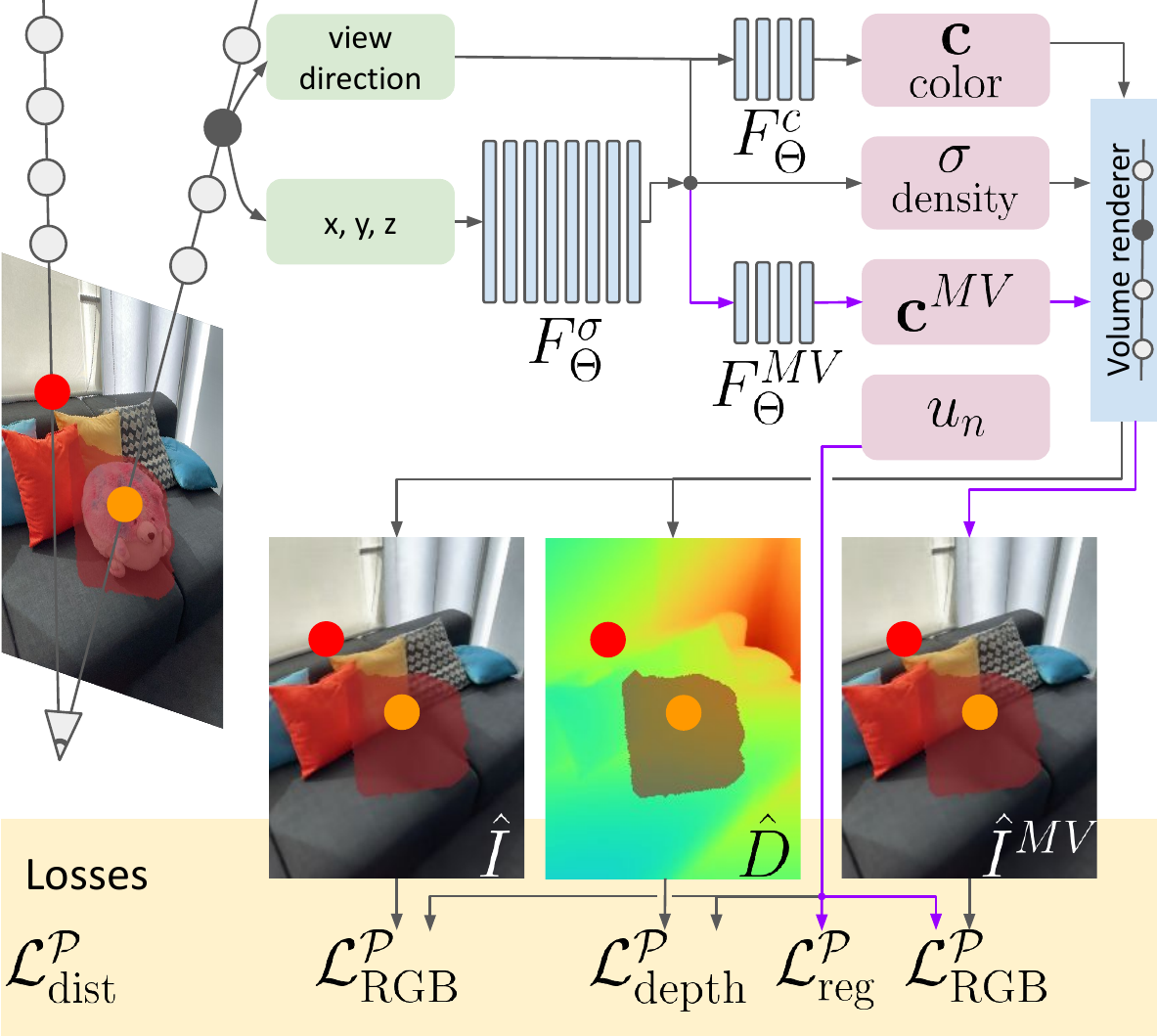} 
    \caption{\paragraph{Our architecture} Our NeRF formulation contains two color heads $F_{\Theta}^c$ and $F_{\Theta}^{MV}$, where  $F_{\Theta}^{MV}$ does not take the view direction as input. This is important to encourage multi-view consistency (see text for details).
    It also includes uncertainty variables $u_n$ that are used to jointly model the scene and the uncertainty of the 2D inpaintings for our automated view selection.
    }
    \label{fig:architecture}
    \vspace{-10pt}
\end{figure}

\section{Method}
\label{sec:method}

We assume we are given an RGB-D sequence with camera poses and intrinsics. 
Depths and poses could be acquired, for example, using a dense structure-from-motion pipeline \cite{schoenberger2016mvs,schonberger2016structure}.
For most of our experiments we capture posed RGB-D sequences directly using Apple's ARKit framework \cite{arkit}, but we also show that we can relax this requirement through use of a multi-view stereo method to estimate depth for an RGB sequence.
Along the way, we also assume access to a per-frame mask of the object to be removed. 
The goal is to learn a NeRF model from this input, which can be used to synthesize consistent novel views, where the per-frame masked region should be plausibly inpainted. An overview of our method is shown in Fig.~\ref{fig:method_overview}.

\subsection{RGB and depth inpainting network}
Our method relies on a 2D single image inpainting method to inpaint each RGB image individually. Furthermore, we also require a depth inpainting network. We use both networks as black-boxes and our approach is agnostic to the method chosen. Future improvements in single image inpainting can be directly translated to improvements to our method.
Given an image $I_n$ and corresponding mask $M_n$, the per-image inpainting algorithm produces a new image $\tilde{I}_n$. Similarly, the depth inpainting algorithm produces an inpainted depth map $\tilde{D}_n$. We show some results for the 2D inpainting network in Figure~\ref{fig:static-inpainting-inconsistent}.

\subsection{Background on NeRFs}
Following the original NeRF paper \cite{mildenhall2020nerf}, we represent the scene as an MLP $F_{\Theta}$
that predicts color $\textbf{c} = [r,g,b]$ and density $\sigma$, for a 5 dimensional input containing $x,y,z$ position and two viewing directions.
The predicted color for pixel $\mathbf{r}$, $\hat{I}_n(\mathbf{r})$, is obtained by volume rendering along its associated ray, so
\begin{equation}
\resizebox{\columnwidth}{!}{$\displaystyle{
    \hat{I}_n(\textbf{r}) = \sum_{i=1}^K \underbrace{T_i (1-\exp{(-\sigma_i \delta_i)})}_{w_i} \textbf{c}_i,
    ~T_i = \exp\left(-\sum_{j=1}^{i-1}\sigma_j \delta_j\right)   
    }$,}
\end{equation}
where $K$ is the number of samples along the ray, $t_i$ is a sample location, $\delta_i = t_{i+1} - t_{i}$ is the distance between adjacent samples, and $w_i$ is the alpha compositing weight which, by construction, sum to less than or equal to 1.

The NeRF loss operates on training images as
\begin{equation}
    \mathcal{L}_\text{RGB} = \sum_{n=1}^{N}\sum_{\mathbf{r} \in \Omega_n}\left\Vert I_n(\mathbf{r}) - \hat{I}_n(\mathbf{r})\right\Vert^{2},
\end{equation}
where $I_n(\mathbf{r})$ is the input RGB value for pixel $\mathbf{r}$, and $\hat{I}_n(\mathbf{r})$ is its predicted color.
$\Omega_n$ indicates the 2D domain of image $n$. 
The parameters of the MLP, $\Theta$, are optimized to minimize this loss.
Similarly to \cite{roessle2022dense}, if input depth is available, then an additional loss can be added,
\begin{equation}
    \resizebox{\columnwidth}{!}{$\displaystyle{
        \mathcal{L}_\text{depth} = \sum_{n=1}^{N}\sum_{\mathbf{r} \in \Omega_n}\left|D_{n}(\mathbf{r}) - \hat{D}_n(\mathbf{r})\right|, ~\text{with~}\hat{D}_n(\mathbf{r}) = \sum_{i=1}^K w_it_i
    }$},
\end{equation}
where $D_{n}(\mathbf{r})$ is the input depth for pixel $\mathbf{r}$, and $\hat{D}_n(\mathbf{r})$ is the corresponding predicted depth.

Finally, a distortion regularizer loss was introduced in \cite{barron2022mipnerf360} to better constrain the NeRF optimization and remove ``floaters''. It encourages the non-zero compositing weights $w_i$ 
to be concentrated in a small region along the ray, so for each pixel $\mathbf{r}$,
\begin{equation}
    l_\text{dist}(\mathbf{r}) = \sum_{i, j}{w_i w_j \left| \frac{t_i + t_{i+1}}{2} - \frac{t_j + t_{j+1}}{2}\right| }+
    \frac{1}{3}\sum_{i}{w_i^2 \delta_i} \label{eq:L_dist_ray}.
\end{equation}

\subsection{Confidence-based view selection}\label{sec:view_selection}
Despite most of the individual inpainted RGB images $\tilde{I}_n$ looking realistic, they still suffer from two issues: 1) some of the inpaintings are incorrect, and 2) despite individual plausibility, they are not multi-view consistent, \ie the same area observed in multiple views is not necessarily completed in a consistent way (Figure~\ref{fig:static-inpainting-inconsistent}).
For this reason, we propose a confidence-based view selection scheme, that automatically chooses which views are used in the NeRF optimization.
We associate to each image $n$ a non-negative uncertainty measure $u_n$. The corresponding per-image confidence, $e^{-u_n}$, is used to re-weight the NeRF losses. This confidence value can be seen as a loss attenuation term, similar to the aleatoric uncertainty prediction term in \cite{kendall2017uncertainties}.

The RGB loss for our model is then set out as
\begin{equation}
    \resizebox{\columnwidth}{!}{$\displaystyle{
        \mathcal{L}_\text{RGB}^{\mathcal{P}}(\hat{I}) = \sum_{n=1}^{N}\sum_{\mathbf{r} \in \Omega_n\setminus  M_n}\left\Vert{I}_n(\mathbf{r}) - \hat{I}_n(\mathbf{r})\right\Vert^{2} 
        +\sum_{n \in \mathcal{P}}e^{-u_n}\sum_{\mathbf{r} \in M_n}\left\Vert\tilde{I}_n(\mathbf{r}) - \hat{I}_n(\mathbf{r})\right\Vert^{2}
    }$\label{eq:rgb_p_loss}
    }
\end{equation}
where the color for pixels $\mathbf{r}$ is supervised by the inpainted image for pixels inside the mask, and by the input RGB image for pixels outside the mask.
Note that the second term of this loss is only computed over a restricted set of images $\mathcal{P}$, where $\mathcal{P} \subseteq \{1,..,N\}$. This is denoted by the superscript $\mathcal{P}$ in $\mathcal{L}_\text{RGB}^{\mathcal{P}}$.
In practice, that means that only some inpainted regions are used in the NeRF optimization. We discuss below how we choose the set $\mathcal{P}$.

We use a similar split into pixels inside and outside the mask for the depth loss, so
\begin{equation}
    \resizebox{\columnwidth}{!}{$\displaystyle{
        \mathcal{L}_\text{depth}^{\mathcal{P}} = \sum_{n=1}^{N}\sum_{\mathbf{r} \in \Omega_n\setminus  M_n}\left|{D}_n(\mathbf{r}) - \hat{D}_n(\mathbf{r})\right| 
        + \sum_{n \in \mathcal{P}}e^{-u_n}\sum_{\mathbf{r} \in M_n} \left|\tilde{D}_n(\mathbf{r}) - \hat{D}_n(\mathbf{r})\right|
    }$}.
\end{equation}

Finally, we include two regularizers. One is on the uncertainty weights $\mathcal{L}_\text{reg}^{\mathcal{P}} = \sum_{n \in \mathcal{P}}u_n$, to prevent a trivial solution where $e^{-u_n}$ is $0$. The other is a distortion regularizer, based on \cite{barron2021mipnerf}, around the loss detailed in \Cref{eq:L_dist_ray}, so 
\begin{align}
    \mathcal{L}_\text{dist}^{\mathcal{P}} &= \sum_{n=1}^{N}\sum_{\mathbf{r} \in \Omega_n\setminus  M_n} l_\text{dist}(\mathbf{r})
    + \sum_{n \in \mathcal{P}}\sum_{\mathbf{r} \in M_n}l_\text{dist}(\mathbf{r}).
\end{align}

\paragraph{View direction and multi-view consistency}
When optimizing the NeRF, we made three observations: a) the multi-view inconsistencies in the inpaintings are modelled by the network using the viewing direction; b) we can enforce multi-view consistency by removing the viewing direction from the input; c) the inconsistencies introduce cloud-like artefacts in the density when not using the viewing direction as input.
To prevent a) and c) and correctly optimize the variables $u_n$ that capture the uncertainty about the inpaintings
$\tilde{I}_n$, we propose: 1) adding an auxiliary network head, $F_{\Theta}^{MV}$, to the NeRF that does not take the viewing direction as input and, 2) stopping the gradient from the color inpainting and $F_{\Theta}^{MV}$ to the density, leaving the uncertainty variable $u_n$ as the only view-dependent input. 
This design forces the model to encode the inconsistencies between inpaintings into the uncertainty prediction while keeping the model consistent across views. 
$F_{\Theta}^{MV}$ has a loss term based on \mbox{Eq.~\ref{eq:rgb_p_loss}}:
$\mathcal{L}_\text{RGB}^{\mathcal{P}}(\hat{I}^{MV})$.
See \Cref{fig:architecture} for an illustration of our architecture.

Our final loss is then $\mathcal{L}^{\mathcal{P}} = \lambda_\text{RGB} \mathcal{L}_\text{RGB}^{\mathcal{P}}(\hat{I}) +
\lambda_\text{RGB} \mathcal{L}_\text{RGB}^{\mathcal{P}}(\hat{I}^{MV}) + \lambda_\text{depth} \mathcal{L}_\text{depth}^{\mathcal{P}} + \lambda_\text{reg} \mathcal{L}_\text{reg}^{\mathcal{P}} + \lambda_\text{dist} \mathcal{L}_\text{dist}^{\mathcal{P}}$, which is optimized over the MLP parameters $\Theta=\left\{ {\Theta}^{\sigma}, {\Theta}^{c}, {\Theta}^{MV}\right\}$  and the uncertainty weights $\mathbf{U}^\mathcal{P} = \{u_n, n \in \mathcal{P}\}$.
The confidence of all images is initialized to $1$, \ie $u_n := 0$.

\paragraph{Iterative refinement}
We use the predicted per-image uncertainty, $u_n$, in an iterative scheme that progressively removes non-confident images from the NeRF optimization, \ie we iteratively update the set $\mathcal{P}$ of images that contribute to the loss in masked regions.
After $K_\text{grad}$ steps of optimizing $\mathcal{L}^\mathcal{P}$, we find the median estimated confidence value $m$.
We then remove from the training set all 2D \emph{inpainted regions} which have associated confidence scores less than $m$.
We then retrain the NeRF with the updated training set, and repeat these steps $K_\text{outer}$ times. Note that images excluded from $\mathcal{P}$ still participate in the optimization, but only for rays in the unmasked regions as they contain valuable information about the scene.
This is summarized in Algorithm~\ref{algo:iterative_refinement}.

\newcommand\mycommfont[1]{\scriptsize\ttfamily{#1}}
\SetCommentSty{mycommfont}
\begin{algorithm}[t]
    \footnotesize
    \SetAlgoLined
    \KwData{\small Input images $I_n$, Inpainted images $\tilde{I}_n$, Depth maps $D_n$, Inpainted depth maps $\tilde{D}_n$, Masks $M_n$}
    \KwResult{\small Trained NeRF model $F_\Theta$ with object removed}
    \tcc{Set of images used for training NeRF is initialized with all images.}
    $\mathcal{P} \gets \{1,...,N\}$ and  $u_n \gets 0,~~n \in \mathcal{P}$ \\
    \For {$i \gets 0$ \KwTo $K_\text{outer}$}
    {  
        $\Theta \gets $ randomly initialized\\
        \For {$j \gets 0$ \KwTo $K_\text{grad}$}
        {
            \tcc{Gradient iterations of NeRF training.}
            $\Theta \gets \Theta - \nabla_{\Theta}\mathcal{L}^\mathcal{P}$ \\
            $\mathbf{U}^{\mathcal{P}} \gets \mathbf{U}^{\mathcal{P}} - \nabla_{\mathbf{U}^\mathcal{P}}\mathcal{L}^{\mathcal{P}}$
        }
        \tcc{Calculate median of confidence values.}
        $m = \text{median}(\{e^{-u_n}, n \in \mathcal{P}\})$ \\
        \tcc{Update $\mathcal{P}$ by removing images with small confidence.}
        \For {$n \in \mathcal{P}$}
        {
            \If{$e^{-u_n} < m$}{
                $\mathcal{P} \gets \mathcal{P} \setminus n $
            }
        }
    }
    \caption{Iterative refinement using confidence based view selection.}
    \label{algo:iterative_refinement}
    
\end{algorithm}

\subsection{Implementation details}

\paragraph{Masking the object to be removed}
Similarly to other inpainting methods, our method requires per-frame masks as input.
Manually annotating each frame with a 2D mask, as done in other inpainting methods \cite{suvorov2022lama, xu2019VideoInpainting, liCvpr22vInpainting}, is time consuming. Instead, we propose to manually annotate a 3D box that contains the object, which only has to be done once per scene.
We reconstruct a 3D point cloud of the scene using the camera poses and the input depth maps. This point cloud can then be imported into a 3D 
visualisation and editing software, such as MeshLab \cite{CignoniCR08}, where we specify a 3D bounding box enclosing the object to be removed.
Alternatively, we could have relied on 2D object segmentation methods, \eg \cite{he2017maskRcnn}, or 3D object bounding box detectors, \eg one of the baselines in \cite{objectron2021}, to mask the object.

\begin{figure}
    \centering
    \footnotesize
    \setlength{\tabcolsep}{1.5pt}
    \begin{tabular}{cc|cc}
    \multicolumn{2}{c|}{Raw masks} & \multicolumn{2}{c}{Refined masks} \\
    \includegraphics[width=0.24\columnwidth]{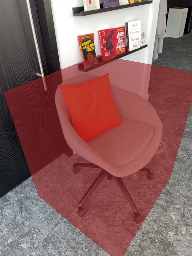} & 
    \includegraphics[width=0.24\columnwidth]{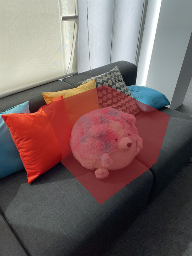} & 
    \includegraphics[width=0.24\columnwidth]{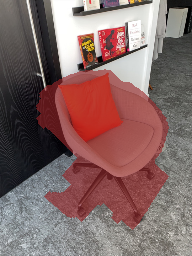} & 
    \includegraphics[width=0.24\columnwidth]{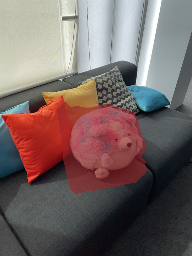} 
    \end{tabular}
    \vspace{-5pt}
    \caption{\textbf{Mask refinement} Our mask refinement leads to smaller masks and therefore higher quality inpainting.}
    \label{fig:mask_refinement}
\end{figure}

\paragraph{Mask refinement} In practice, we observe that masks obtained from the annotated 3D bounding boxes can be quite coarse and include large parts of the background. Since large masks have a negative effect on the inpainting quality, we propose a mask refinement step to obtain masks which are tighter around the object. This step is not required for tight input masks.
Intuitively, this mask refinement step removes parts of the 3D bounding box that are empty space. 
We start by taking all points in the reconstructed 3D point cloud that are inside the 3D bounding box. 
The refined mask is then obtained by rendering these points into each image and performing a simple comparison with the depth map to check occlusions in the current image.
The resulting mask is cleaned up by dilating any pixel leaks caused by sensor noise using binary dilation and erosion.
The effect of our mask refinement step can be seen in Figure~\ref{fig:mask_refinement}.

\input{main_results_table}

\paragraph{Inpainting network}
We used \cite{suvorov2022lama} for inpainting both RGB and depth. 
The inpainting of RGB images and depth maps is done independently and we use the reference network provided by the authors of \cite{suvorov2022lama} for both. 
Our depth maps are preprocessed by clipping to \SI{5}{\m} and linearly mapping depths in $\left[\SI{0}{\m}, \SI{5}{\m}\right]$ to pixel values of $\left[0, 255\right]$. This matrix is duplicated to make a $H\times W \times 3$ tensor for input to \cite{suvorov2022lama}.
We observed empirically that this approach provided good results, but an inpainting method specific for depth maps could improve over this baseline.

\paragraph{NeRF estimation}
The implementation of our method is built upon \cite{Niemeyer2021Regnerf} and \cite{barron2021mipnerf}.
The shared MLP $F_{\Theta}^{\sigma}$ 
consists of eight 256-wide layers while the branches $F_{\Theta}^{c}$ and $F_{\Theta}^{MV}$ consist of four 128-wide layers each. 
The final output is activated using \emph{softplus} for density, ReLU for the uncertainty, and sigmoid for the color channel.
We weight the terms in the loss function with
$\lambda_\text{RGB} = \lambda_\text{depth} = \lambda_\text{dist} = 1$, and $\lambda_\text{reg} = 0.005$.
We optimize using Adam with an initial learning rate of $l_r = 0.0005$, and use the scheduler from \cite{Niemeyer2021Regnerf}.
We do a filtering step, where we remove low confidence images every $K_\text{grad} = 50,000$ steps, resulting in $K_\text{outer} = 4$ filtering steps.
More details concerning implementation are in the supplementary material.

\section{Experiments}

\subsection{Datasets}

While previous approaches have tackled static object removal from videos, no standard dataset/metrics to evaluate these systems has been proposed, to our knowledge.
This work introduces an RGB-D dataset of real scenes, designed to evaluate the quality of object removal. Our dataset will be made public and has two variants, which are used differently when benchmarking:

\paragraph{Ours --- Real objects} This dataset comprises 17 scenes, which are a mixture of indoors/outdoors with landscape/portrait orientation, and focus on a small area with at least one object, with one of them being the labeled object of interest. They vary in difficulty in terms of background texture, size of the object, and complexity of scene geometry.
Please see supplementary material for a visualization of the full dataset.
For each scene, we collected two sequences, one with and the other without the object that we want to remove. The sequences are collected using ARKit~\cite{arkit} on an iPhone 12 Pro with Lidar, and contain RGB-D images and poses. The RGB images are resized to $192\times256$, the same resolution as the depth maps.
Both sequences for the same scene use the same camera coordinate system.
The sequences vary in length, from 194 to 693 frames.

As previously described, the masks are obtained by annotating a 3D bounding box of the object of interest and are refined for all of the scenes.
For each scene, we use the sequence with the object and corresponding masks for training the NeRF model, and the sequence without the object for 
testing.
The use of real objects makes it easier to evaluate how the systems deal with real shadows and reflections, as well as novel view synthesis.

\paragraph{Ours --- Synthetic objects} 
Most video and image inpainting methods, \eg \cite{suvorov2022lama, liCvpr22vInpainting}, do not perform novel view synthesis, meaning such methods cannot be fairly evaluated on our `Real objects' dataset. 
We therefore introduce a separate synthetically-augmented variant of our dataset.
This uses the same scenes as the real objects dataset, but we only use the sequence that \emph{doesn't} contain the object. 
We then manually position a 3D object mesh from ShapeNet \cite{shapenet2015} in each scene. The object is placed so that it has a plausible location and size, \eg a laptop on a table. 
The masks are obtained by projecting the mesh into the input images, which is the only use we make of the 3D object mesh. 
Please see supplementary material for a visualisation of the full synthetic dataset.
For this synthetic dataset, following \eg \cite{barron2022mipnerf360}, we use every 8th frame for testing and the rest of the frames for training the NeRF model.

\paragraph{ARKitScenes}
We further validate our approach qualitatively on ARKitScenes \cite{arkitscenes}.
This is an RGB-D dataset of 1,661 scenes, where depth was captured via iPhone Lidar.

\subsection{Metrics}
To evaluate the object removal and inpainting quality, we compare our system's output image against the ground truth image, for each test image in the dataset.
All metrics in the paper are only computed inside the masked region. Metrics for the full image are provided in the supplementary material.
We use the three standard metrics for NeRF evaluation~\cite{mildenhall2020nerf}:
PSNR \cite{hore2010image}, SSIM \cite{wang2004image} and LPIPS \cite{zhang2018perceptual}.
To evaluate the geometric completion, we compute the $L_1$ and $L_2$ error between the rendered and the ground-truth depth maps inside masked regions.
The metrics are averaged over all frames of a sequence, and then averaged over all sequences.

\begin{figure}
    \centering
    \footnotesize
    \setlength{\tabcolsep}{1pt}
    \newcommand{\imwidth}{0.13\textwidth}
    \begin{tabular}{cccc}
    \rotatebox{90}{\phantom{xxxxxxxxx}Input}& 
    \includegraphics[width=0.3\columnwidth]{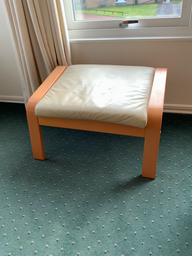} & 
    \includegraphics[width=0.3\columnwidth]{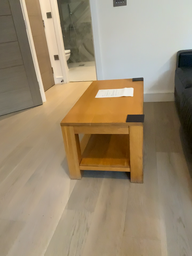} & 
    \includegraphics[width=0.3\columnwidth]{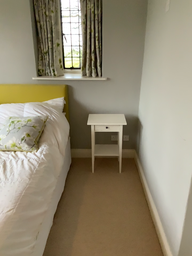} \\
    \rotatebox{90}{\phantom{xxxxxx}Rendering} & 
    \includegraphics[width=0.3\columnwidth]{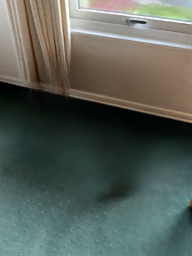} & 
    \includegraphics[width=0.3\columnwidth]{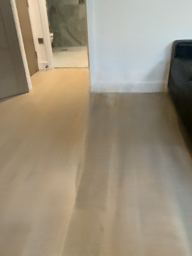} & 
    \includegraphics[width=0.3\columnwidth]{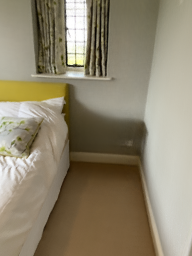} \\
    \end{tabular}
    \vspace{-6pt}
    \caption{\textbf{Results on ARKitScenes \cite{arkitscenes}}. We can successfully remove objects from casually captured sequences in indoor scenes.}
    \label{fig:arkitscenes_qual}
    \vspace{-5pt}
\end{figure}

\subsection{Comparison with baselines}

In Table~\ref{tab:baseline}, we compare our approach with alternative methods for object removal, with a focus on methods that use an underlying NeRF representation.

\paragraph{Image and video inpainting baselines} We compare with two state-of-the-art methods for image \method{LaMa~\cite{suvorov2022lama}} and video inpainting \method{E$^2$FGVI~\cite{liCvpr22vInpainting}}. 
In both cases, we use their reference implementation and provided trained network.
Neither of these methods allows novel view synthesis, so they are only evaluated on the synthetic objects dataset.

\paragraph{NeRF-based ablations} We compare different ways of training a baseline NeRF model. The \method{Masked NeRF} baseline
corresponds to training a NeRF using the full input RGB-D data, but pixels and depths in the masked regions are ignored in the NeRF losses.
\method{Inpainted Images} is a NeRF trained with all inpainted images, but not using the inpainted depth maps, while \method{Inpainted Images + Inpainted Depth} uses all the inpainted images and inpainted depth maps. This baseline corresponds to \method{All views} in Table \ref{tab:ablation_view_selection}.

\paragraph{3D scene completion} We compare with several published works for 3D scene completion. Note that none of these baselines specifically targets inpainting, so results should be viewed with this in mind. For all of them, we use their publicly available implementation.
\method{PixelSynth \cite{rockwell2021pixelsynth}} and \method{CompNVS \cite{li2022compnvs}} were both proposed for scene \emph{outpainting}. Given one or a few frames of a scene their goal is to complete the scene to enable novel view synthesis. Both of these methods rely on a generative model of indoor scenes and neither requires test-time optimization. Both methods are adapted to use our masks as input. We show qualitative results for \method{CompNVS} in the supplementary materials.
\method{Object compositional NeRF \cite{yang2021objectnerf}} enables objects in NeRFs to be edited via user-editing of pose transformations. We adapt their code for object \emph{removal} by setting a transformation that moves the object outside the camera's field of view.

\paragraph{Our method} We compare two different versions of our method. \method{Ours} is our proposed approach, which uses the method described in Section \ref{sec:method}.
\method{Ours -- depth from \cite{sayed2022simplerecon}} is our method, where the depth maps are obtained using a state-of-the art multi-view depth prediction method \cite{sayed2022simplerecon} to show that we do not necessarily rely on sensor depth.

As shown in Table~\ref{tab:baseline}, our method is superior to other novel-view synthesis baselines across most appearance and depth metrics.
Moreover, as opposed to the single image inpainting \method{LaMa~\cite{suvorov2022lama}}, our method is close to multi-view consistent, significantly reducing inter-frame flickering. 
To scrutinize the flickering, we refer to the supplementary video.

\begin{figure*}[t]
    \centering
    \footnotesize
    \setlength{\tabcolsep}{1pt}
    \newcommand{\imwidth}{0.13\textwidth}
    \begin{tabular}{ccccccc}
        Input & Masked NeRF$\dagger$ & Inpainted NeRF$\dagger$ & Obj.~comp.~\cite{yang2021objectnerf} & PixelSynth \cite{rockwell2021pixelsynth} & Ours & Ground-truth\\

        \includegraphics[width=\imwidth]{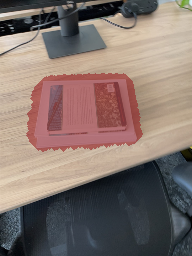} &
        \includegraphics[width=\imwidth]{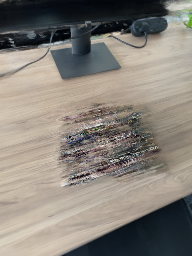} &
        \includegraphics[width=\imwidth]{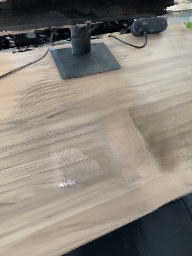} &
        \includegraphics[width=\imwidth]{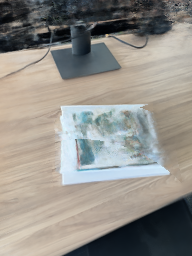} &
        \includegraphics[width=\imwidth]{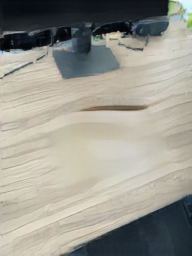} &
        \includegraphics[width=\imwidth]{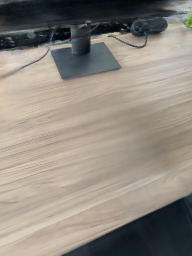} &
        \includegraphics[width=\imwidth]{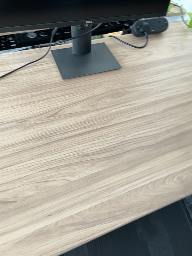} \\

        \includegraphics[width=\imwidth]{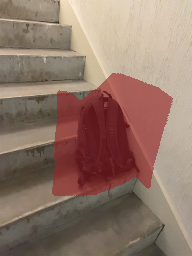} &
        \includegraphics[width=\imwidth]{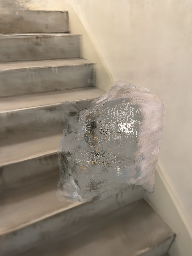} &
        \includegraphics[width=\imwidth]{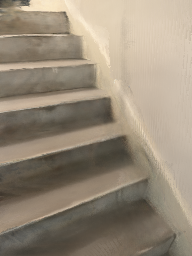} &
        \includegraphics[width=\imwidth]{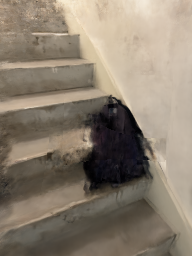} &
        \includegraphics[width=\imwidth]{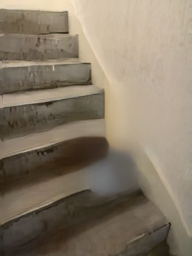} &
        \includegraphics[width=\imwidth]{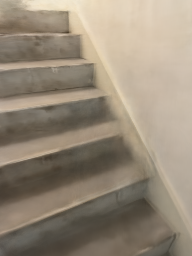} &
        \includegraphics[width=\imwidth]{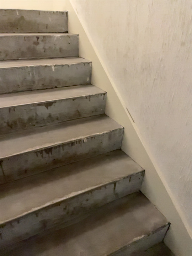} \\

        \includegraphics[width=\imwidth]{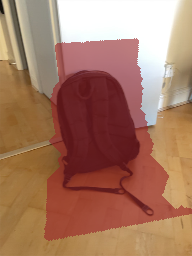} &
        \includegraphics[width=\imwidth]{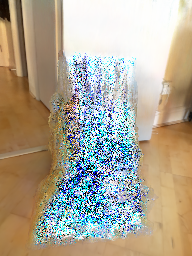} &
        \includegraphics[width=\imwidth]{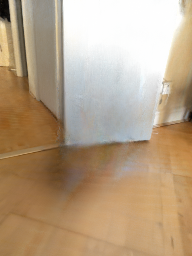} &
        \includegraphics[width=\imwidth]{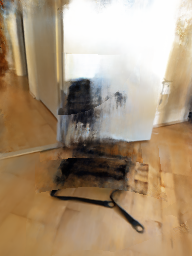} &
        \includegraphics[width=\imwidth]{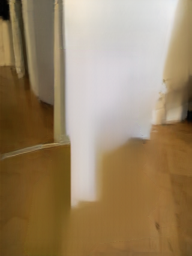} &
        \includegraphics[width=\imwidth]{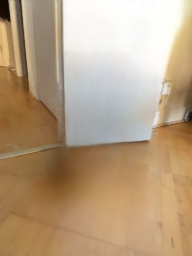} &
        \includegraphics[width=\imwidth]{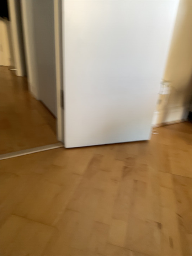} \\

        \includegraphics[width=\imwidth]{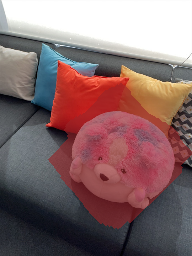} &
        \includegraphics[width=\imwidth]{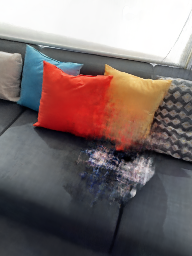} &
        \includegraphics[width=\imwidth]{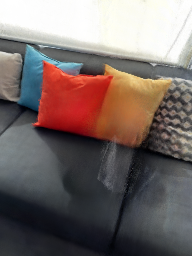} &
        \includegraphics[width=\imwidth]{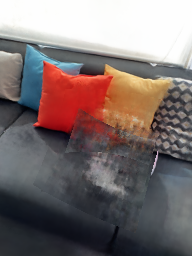} &
        \includegraphics[width=\imwidth]{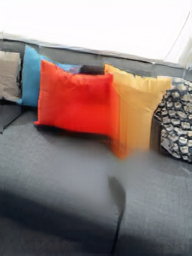} &
        \includegraphics[width=\imwidth]{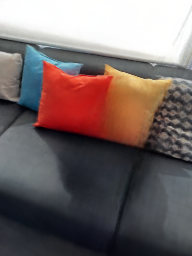} &
        \includegraphics[width=\imwidth]{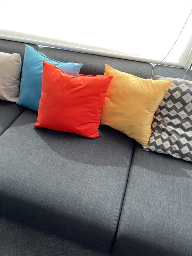} \\

         Input & Masked NeRF$\dagger$ & Inpainted NeRF$\dagger$ & Obj.~comp.~\cite{yang2021objectnerf} & PixelSynth \cite{rockwell2021pixelsynth} & Ours & Ground-truth\\
    \end{tabular}
    \vspace{-6pt}
    \caption{\textbf{Qualitative comparisons with baseline.} Our method significantly improves over 3D scene completion baselines for the task of removing objects from a scene using 3D inpainting. Further, it mitigates artefacts and stabilizes convergence compared to the inpainting baseline without automatic view selection.
    $\dagger$: We also compare with two NeRF based baselines: \method{Masked NeRF} and \method{Inpainted NeRF}.
    }
    \label{fig:qualitative_depths}
    \vspace{-10pt}
\end{figure*}

\paragraph{Qualitative comparison}
In Figure~\ref{fig:qualitative_depths}, we show that the proposed method successfully removes the selected object compared to the baselines. 
While \method{Masked NeRF} fails to complete large holes and \method{Inpainted NeRF} suffers from bad inpaintings in the training set, our method can leverage the 2D inpaintings, while avoiding integrating artefacts by removing those input frames. In contrast to \method{Object compositional NeRF \cite{yang2021objectnerf}}, leveraging the inpaintings also helps to mitigate the appearance of artefacts below the object's surface.
Compared to \cite{yang2021objectnerf} and \cite{rockwell2021pixelsynth}, our method is better able to generate plausible scene completions. We also show results from the ARKitScenes dataset in Figure~\ref{fig:arkitscenes_qual}, and please also see our supplementary video.

\begin{table}%
    \centering
    \resizebox{1.0\columnwidth}{!}{
    \begin{tabular}{lcccccc}
        \toprule
        Method &  \# of views & PSNR$\uparrow$ & SSIM$\uparrow$ & LPIPS$\downarrow$ & $L_1\downarrow$ & $L_2\downarrow$
        \\
        \midrule
        All views	&	82 - 382	&	27.568	&	0.898	&	0.094	&	0.231	&	0.318	\\
        1/10th	&	8 - 38	&	27.098	&	0.900	&	0.079	&	0.202	&	0.291	\\
        1/50th	&	1 - 7	&	26.718	&	0.893	&	0.087	&	0.229	&	0.309	\\
        Single view	&	1	&	26.232	&	0.892	&	0.079	&	0.133	&	0.198	\\
        Ours	&	10 - 185	&	\textbf{29.437}	&	\textbf{0.916}	&	\textbf{0.078}	&	\textbf{0.069} &	\textbf{0.096}	\\
        \bottomrule
    \end{tabular}
    }
    \vspace{-5pt}
    \caption{\textbf{Ablation on view selection methods.} We validate our view selection formulation by comparing to alternative approaches. Ours consistently produces better performing models.}
    \label{tab:ablation_view_selection}
\end{table}

\subsection{View selection ablation}
Here we validate that our view selection procedure from Section \ref{sec:view_selection} contributes to improved results.
We compare our method with different  view selection strategies in Table~\ref{tab:ablation_view_selection} on our synthetic object dataset.
\method{All views} uses all the inpainting views when training the NeRF model. The other baselines use a subset of views to train the NeRF model, spaced at regular intervals: every 10th frame for \method{1/10th}; every 50th frame for \method{1/50th} and a single middle frame for \method{Single view}. The number of views used for each sequence varies depending on the length of the sequence.
We outperform these baselines, suggesting that our proposed strategy for view selection is effective in choosing a good set of views to include in the NeRF optimization.

\subsection{Limitations}
Our method is upper bounded by the performance of the 2D inpainting method. 
When the masks are too large along the entire trajectory, the 2D inpainting fails and no realistic views can be selected.
Our renderings sometimes suffer from blurring, caused by flickering of high-frequency textures in the 2D inpaintings.
Furthermore, cast shadows or reflections of the object are not handled well.
We leave tackling these challenges for future work.

\begin{figure}
    \centering
    \footnotesize
    \setlength{\tabcolsep}{1pt}
    \newcommand{\imwidth}{0.13\textwidth}
    \begin{tabular}{cccc}
    Input & Rendering & Input & Rendering \\
    \includegraphics[width=0.24\columnwidth]{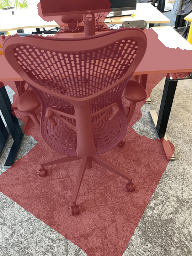} & 
    \includegraphics[width=0.24\columnwidth]{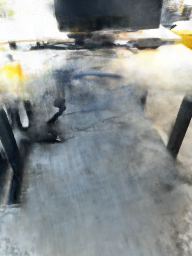} & 
    \includegraphics[width=0.24\columnwidth]{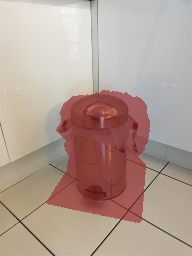} & 
    \includegraphics[width=0.24\columnwidth]{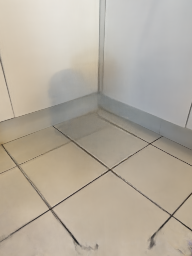} \\
    \end{tabular}
    \vspace{-6pt}
    \caption{\textbf{Failure cases and limitations.} Our method can not recover when the 2D inpainting method fails all the frames, for example when the mask and covers a large part of the image. Further, our method keeps the shadows of the removed object, if they are not included in the object mask.}
    \vspace{-6pt}
\end{figure}

\section{Conclusion}

We have presented a framework to train neural radiance fields, where objects are plausibly removed from the output renderings.
Our method draws upon existing work in 2D inpainting, and introduces an automatic confidence-based view selection scheme to select single-view inpaintings with multi-view consistency.
We experimentally validated that our proposed method improves novel-view synthesis from 3D inpainted scenes compared to existing work.
We have also introduced a dataset for evaluating this work, which sets a benchmark for other researchers in the field.
\noindent
\begin{minipage}{\columnwidth}
	\vspace{-1mm}
	\footnotesize
	\noindent
	\textbf{Acknowledgments.}~
	We thank Jamie Watson for fruitful discussions about the project, Paul-Edouard Sarlin for the constructive feedback on an early draft, and Zuoyue Li for kindly running CompNVS on our dataset.
    \vspace{-1mm}
\end{minipage}

{\small
\bibliographystyle{ieee_fullname}
\bibliography{ObjectRemoval}
}

\end{document}

%% file: macros.tex
\def\ie{{\it i.e.,\ }}
\def\etal{{et~al.}}

\def\eg{{\it e.g.,\ }}

\usepackage{textcomp}

%% file: main_results_table.tex
\begin{table*}[t]
    \centering
    \footnotesize
    \resizebox{1.0\textwidth}{!}{
        \begin{tabular}{l ccccc ccccc}
            \toprule
            ~ & \multicolumn{5}{c}{\textbf{Synthetic objects - Masked}} & \multicolumn{5}{c}{\textbf{Real objects - Masked}} 
            \\
            \cmidrule(lr){2-6}
            \cmidrule(lr){7-11}
            ~ & PSNR$\uparrow$ & SSIM$\uparrow$ & LPIPS$\downarrow$ & Depth $L_1\downarrow$ & Depth $L_2\downarrow$
              & PSNR$\uparrow$ & SSIM$\uparrow$ & LPIPS$\downarrow$ & Depth $L_1\downarrow$ & Depth $L_2\downarrow$
            \\
            \midrule \hspace{-8pt} \textbf{No novel view synthesis}	&		&		&		&		&		&		&		&		&		&		\\
LaMa \cite{suvorov2022lama}$^{\dagger}$	&	27.999	&	0.898	&	\textbf{0.060}	&	\underline{0.070}	&	\textbf{0.075}	&	-	&	-	&	-	&	-	&	-	\\
E$^2$FGVI \cite{liCvpr22vInpainting}$^{\dagger\ddagger}$	&	24.568	&	0.874	&	0.102	&	-	&	-	&	-	&	-	&	-	&	-	&	-	\\
\midrule \hspace{-8pt} \textbf{NeRF based ablations}	&		&		&		&		&		&		&		&		&		&		\\
Masked NeRF	&	26.126	&	0.882	&	0.093	&	0.084	&	0.105	&	21.644	&	0.815	&	0.142	&	\underline{0.096}	&	\underline{0.054}	\\
Inpainted NeRF	&	\underline{28.760}	&	\underline{0.905}	&	0.086	&	0.278	&	0.400	&	23.705	&	0.848	&	0.134	&	0.145	&	0.121	\\
Inpainted NeRF + inpainted Depth	&	27.568	&	0.898	&	0.094	&	0.231	&	0.318	&	23.652	&	0.844	&	0.136	&	0.313	&	0.387	\\
\midrule \hspace{-8pt} \textbf{3D scene completion}	&		&		&		&		&		&		&		&		&		&		\\
PixelSynth \cite{rockwell2021pixelsynth}$^\ddagger$	&	25.481	&	0.887	&	0.116	&	-	&	-	&	\textbf{25.438}	&	0.851	&	0.152	&	-	&	-	\\
CompNVS \cite{li2022compnvs}	&	17.389	&	0.823	&	0.171	&	1.697	&	3.641	&	-	&	-	& -	&	-	&	-	\\
Object compositional NeRF \cite{yang2021objectnerf}$^{*}$	&	-	&	-	&	-	&	-	&	-	&	21.757	&	0.836	&	0.134	&	0.312	&	0.341	\\
\midrule \hspace{-8pt} \textbf{Our method}	&		&		&		&		&		&		&		&		&		&		\\
Ours	&	\textbf{29.437}	&	\textbf{0.916}	&	\underline{0.078}	&	\textbf{0.069}	&	\underline{0.096}	&	\underline{25.271}	&	\textbf{0.859}	&	\textbf{0.125}	&	\textbf{0.071}	&	\textbf{0.044}	\\
Ours -- depth predicted using \cite{sayed2022simplerecon}	&	26.540	&	0.895	&	0.087	&	0.112	&	0.118	&	25.010	&	\underline{0.856}	&	\underline{0.128}	&	0.142	&	0.140	\\
            \bottomrule
        \end{tabular}
        }
    \vspace{-8pt}
    \caption{\textbf{Comparison with baselines and state of the art methods.} Our method is either \textbf{best} or \underline{second-best} compared to other novel-view synthesis baselines in inpainting the missing regions of the scene, by propagating multi-view information and leveraging 2D inpainting information. Notes: $^{\dagger}$These methods can't be evaluated on the proposed real dataset as they do not synthesize novel views. $^{\ddagger}$These methods do not produce depth maps. $^{*}$\cite{yang2021objectnerf} requires the actual object therefore it can't be evaluated on the proposed synthetic dataset.
    }
    \label{tab:baseline}
    \vspace{-3pt}
\end{table*}

%% file: main.bbl
\begin{thebibliography}{10}\itemsep=-1pt

\bibitem{objectron2021}
Adel Ahmadyan, Liangkai Zhang, Artsiom Ablavatski, Jianing Wei, and Matthias
  Grundmann.
\newblock Objectron: A large scale dataset of object-centric videos in the wild
  with pose annotations.
\newblock In {\em CVPR}, 2021.

\bibitem{arkit}
Apple.
\newblock {ARKit}.
\newblock {Accessed: 14 October 2022}.

\bibitem{barron2021mipnerf}
Jonathan~T. Barron, Ben Mildenhall, Matthew Tancik, Peter Hedman, Ricardo
  Martin-Brualla, and Pratul~P. Srinivasan.
\newblock {Mip-NeRF}: A multiscale representation for anti-aliasing neural
  radiance fields.
\newblock In {\em ICCV}, 2021.

\bibitem{barron2022mipnerf360}
Jonathan~T. Barron, Ben Mildenhall, Dor Verbin, Pratul~P. Srinivasan, and Peter
  Hedman.
\newblock {Mip-NeRF} 360: Unbounded anti-aliased neural radiance fields.
\newblock In {\em CVPR}, 2022.

\bibitem{arkitscenes}
Gilad Baruch, Zhuoyuan Chen, Afshin Dehghan, Tal Dimry, Yuri Feigin, Peter Fu,
  Thomas Gebauer, Brandon Joffe, Daniel Kurz, Arik Schwartz, and Elad Shulman.
\newblock {ARK}itscenes - a diverse real-world dataset for {3D} indoor scene
  understanding using mobile {RGB-D} data.
\newblock In {\em NeurIPS}, 2021.

\bibitem{bei2020dynamic}
Borna Be{\v{s}}i{\'c} and Abhinav Valada.
\newblock Dynamic object removal and spatio-temporal {RGB-D} inpainting via
  geometry-aware adversarial learning.
\newblock {\em IEEE Transactions on Intelligent Vehicles}, 7(2), 2022.

\bibitem{shapenet2015}
Angel~X. Chang, Thomas Funkhouser, Leonidas Guibas, Pat Hanrahan, Qixing Huang,
  Zimo Li, Silvio Savarese, Manolis Savva, Shuran Song, Hao Su, Jianxiong Xiao,
  Li Yi, and Fisher Yu.
\newblock {ShapeNet: An Information-Rich 3D Model Repository}.
\newblock Technical Report arXiv:1512.03012 [cs.GR], Stanford University ---
  Princeton University --- Toyota Technological Institute at Chicago, 2015.

\bibitem{chen2021mvsnerf}
Anpei Chen, Zexiang Xu, Fuqiang Zhao, Xiaoshuai Zhang, Fanbo Xiang, Jingyi Yu,
  and Hao Su.
\newblock {MVSNeRF}: Fast generalizable radiance field reconstruction from
  multi-view stereo.
\newblock In {\em ICCV}, 2021.

\bibitem{chibane2020implicit}
Julian Chibane, Thiemo Alldieck, and Gerard Pons-Moll.
\newblock Implicit functions in feature space for {3D} shape reconstruction and
  completion.
\newblock In {\em CVPR}, 2020.

\bibitem{CignoniCR08}
Paolo Cignoni, Massimiliano Corsini, and Guido Ranzuglia.
\newblock {MeshLab}: an open-source {3D} mesh processing system.
\newblock {\em ERCIM News}, (73), 2008.

\bibitem{davis2012unstructured}
Abe Davis, Marc Levoy, and Fredo Durand.
\newblock Unstructured light fields.
\newblock In {\em Computer Graphics Forum}, volume~31, pages 305--314. Wiley
  Online Library, 2012.

\bibitem{kangle2021dsnerf}
Kangle Deng, Andrew Liu, Jun-Yan Zhu, and Deva Ramanan.
\newblock Depth-supervised {NeRF}: Fewer views and faster training for free.
\newblock In {\em CVPR}, June 2022.

\bibitem{devries2021unconstrained}
Terrance DeVries, Miguel~Angel Bautista, Nitish Srivastava, Graham~W Taylor,
  and Joshua~M Susskind.
\newblock Unconstrained scene generation with locally conditioned radiance
  fields.
\newblock In {\em ICCV}, 2021.

\bibitem{fujii2020rgb}
Ryo Fujii, Ryo Hachiuma, and Hideo Saito.
\newblock {RGB-D} image inpainting using generative adversarial network with a
  late fusion approach.
\newblock In {\em International Conference on Augmented Reality, Virtual
  Reality and Computer Graphics}, pages 440--451. Springer, 2020.

\bibitem{gao2020flow}
Chen Gao, Ayush Saraf, Jia-Bin Huang, and Johannes Kopf.
\newblock Flow-edge guided video completion.
\newblock In {\em ECCV}, 2020.

\bibitem{he2017maskRcnn}
Kaiming He, Georgia Gkioxari, Piotr Doll{\'a}r, and Ross Girshick.
\newblock Mask {R-CNN}.
\newblock In {\em ICCV}, 2017.

\bibitem{hedman2018deep}
Peter Hedman, Julien Philip, True Price, Jan-Michael Frahm, George Drettakis,
  and Gabriel Brostow.
\newblock Deep blending for free-viewpoint image-based rendering.
\newblock {\em ACM Transactions on Graphics (TOG)}, 37(6):1--15, 2018.

\bibitem{herling2010advanced}
Jan Herling and Wolfgang Broll.
\newblock Advanced self-contained object removal for realizing real-time
  diminished reality in unconstrained environments.
\newblock In {\em ISMAR}, 2010.

\bibitem{hore2010image}
Alain Hore and Djemel Ziou.
\newblock Image quality metrics: {PSNR} vs. {SSIM}.
\newblock In {\em ICPR}. IEEE, 2010.

\bibitem{hsu2021moving}
Ching-Yu Hsu, Cheng Sun, and Hwann-Tzong Chen.
\newblock Moving in a 360 world: Synthesizing panoramic parallaxes from a
  single panorama.
\newblock {\em arXiv preprint arXiv:2106.10859}, 2021.

\bibitem{huang2016temporally}
Jia-Bin Huang, Sing~Bing Kang, Narendra Ahuja, and Johannes Kopf.
\newblock Temporally coherent completion of dynamic video.
\newblock {\em ACM Transactions on Graphics (TOG)}, 35(6), 2016.

\bibitem{iizuka2017globally}
Satoshi Iizuka, Edgar Simo-Serra, and Hiroshi Ishikawa.
\newblock Globally and locally consistent image completion.
\newblock {\em ACM Transactions on Graphics (ToG)}, 2017.

\bibitem{kendall2017uncertainties}
Alex Kendall and Yarin Gal.
\newblock What uncertainties do we need in bayesian deep learning for computer
  vision?
\newblock {\em NeurIPS}, 30, 2017.

\bibitem{kim2021generative}
Joohyung Kim, Janghun Hyeon, and Nakju Doh.
\newblock Generative multiview inpainting for object removal in large indoor
  spaces.
\newblock {\em International Journal of Advanced Robotic Systems}, 2021.

\bibitem{kim2022infonerf}
Mijeong Kim, Seonguk Seo, and Bohyung Han.
\newblock Infonerf: Ray entropy minimization for few-shot neural volume
  rendering.
\newblock In {\em CVPR}, 2022.

\bibitem{kobayashi2022decomposing}
Sosuke Kobayashi, Eiichi Matsumoto, and Vincent Sitzmann.
\newblock Decomposing {NeRF} for editing via feature field distillation.
\newblock In {\em NeurIPS}, 2022.

\bibitem{koh2021pathdreamer}
Jing~Yu Koh, Honglak Lee, Yinfei Yang, Jason Baldridge, and Peter Anderson.
\newblock Pathdreamer: A world model for indoor navigation.
\newblock In {\em ICCV}, 2021.

\bibitem{lepetit2001intuitive}
Vincent Lepetit, Marie-Odile Berger, and LORIA-INRIA Lorraine.
\newblock An intuitive tool for outlining objects in video sequences:
  Applications to augmented and diminished reality.
\newblock In {\em ISMAR}, pages 159--160, 2001.

\bibitem{li2022compnvs}
Zuoyue Li, Tianxing Fang, Zhenqiang Li, Zhaopeng Cui, Yoichi Sato, Marc
  Pollefeys, and Martin~R. Oswald.
\newblock {CompNVS}: Novel view synthesis with scene completion.
\newblock In {\em ECCV}, 2022.

\bibitem{liCvpr22vInpainting}
Zhen Li, Cheng-Ze Lu, Jianhua Qin, Chun-Le Guo, and Ming-Ming Cheng.
\newblock Towards an end-to-end framework for flow-guided video inpainting.
\newblock In {\em CVPR}, 2022.

\bibitem{liao2020dvi}
Miao Liao, Feixiang Lu, Dingfu Zhou, Sibo Zhang, Wei Li, and Ruigang Yang.
\newblock {DVI}: Depth guided video inpainting for autonomous driving.
\newblock In {\em ECCV}, 2020.

\bibitem{liu2018image}
Guilin Liu, Fitsum~A Reda, Kevin~J Shih, Ting-Chun Wang, Andrew Tao, and Bryan
  Catanzaro.
\newblock Image inpainting for irregular holes using partial convolutions.
\newblock In {\em ECCV}, 2018.

\bibitem{liu2019learning}
Shichen Liu, Shunsuke Saito, Weikai Chen, and Hao Li.
\newblock Learning to infer implicit surfaces without {3D} supervision.
\newblock {\em arXiv:1911.00767}, 2019.

\bibitem{liu2021editing}
Steven Liu, Xiuming Zhang, Zhoutong Zhang, Richard Zhang, Jun-Yan Zhu, and
  Bryan Russell.
\newblock Editing conditional radiance fields.
\newblock In {\em ICCV}, 2021.

\bibitem{liu2019liquid}
Wen Liu, Zhixin Piao, Jie Min, Wenhan Luo, Lin Ma, and Shenghua Gao.
\newblock Liquid warping {GAN}: A unified framework for human motion imitation,
  appearance transfer and novel view synthesis.
\newblock In {\em ICCV}, 2019.

\bibitem{liu2020learning}
Yu-Lun Liu, Wei-Sheng Lai, Ming-Hsuan Yang, Yung-Yu Chuang, and Jia-Bin Huang.
\newblock Learning to see through obstructions.
\newblock In {\em CVPR}, 2020.

\bibitem{liu2021learning}
Yu-Lun Liu, Wei-Sheng Lai, Ming-Hsuan Yang, Yung-Yu Chuang, and Jia-Bin Huang.
\newblock Learning to see through obstructions with layered decomposition.
\newblock {\em IEEE Transactions on Pattern Analysis and Machine Intelligence},
  44(11), 2021.

\bibitem{Long2022SparseNeuS}
Xiaoxiao Long, Cheng Lin, Peng Wang, Taku Komura, and Wenping Wang.
\newblock {SparseNeuS}: Fast generalizable neural surface reconstruction from
  sparse views.
\newblock In {\em ECCV}, 2022.

\bibitem{lugmayr2022repaint}
Andreas Lugmayr, Martin Danelljan, Andres Romero, Fisher Yu, Radu Timofte, and
  Luc Van~Gool.
\newblock Repaint: Inpainting using denoising diffusion probabilistic models.
\newblock In {\em CVPR}, 2022.

\bibitem{mescheder2019occupancy}
Lars Mescheder, Michael Oechsle, Michael Niemeyer, Sebastian Nowozin, and
  Andreas Geiger.
\newblock Occupancy networks: Learning {3D} reconstruction in function space.
\newblock In {\em CVPR}, 2019.

\bibitem{mildenhall2020nerf}
Ben Mildenhall, Pratul~P. Srinivasan, Matthew Tancik, Jonathan~T. Barron, Ravi
  Ramamoorthi, and Ren Ng.
\newblock {NeRF}: Representing scenes as neural radiance fields for view
  synthesis.
\newblock In {\em ECCV}, 2020.

\bibitem{moreau2022lens}
Arthur Moreau, Nathan Piasco, Dzmitry Tsishkou, Bogdan Stanciulescu, and Arnaud
  de La~Fortelle.
\newblock {LENS: Localization enhanced by NeRF synthesis}.
\newblock In {\em Conference on Robot Learning}, 2022.

\bibitem{mori2020inpaintfusion}
Shohei Mori, Okan Erat, Wolfgang Broll, Hideo Saito, Dieter Schmalstieg, and
  Denis Kalkofen.
\newblock {InpaintFusion}: incremental {RGB-D} inpainting for {3D} scenes.
\newblock {\em IEEE TVCG}, 2020.

\bibitem{Mori2022InpaintFusion}
Shohei Mori, Dieter Schmalstieg, and Denis Kalkofen.
\newblock Good keyframes to inpaint.
\newblock {\em IEEE Transactions on Visualization and Computer Graphics}, 2022.

\bibitem{muller2022instant}
Thomas M{\"u}ller, Alex Evans, Christoph Schied, Marco Foco, Andr{\'a}s
  B{\'o}dis-Szomor{\'u}, Isaac Deutsch, Michael Shelley, and Alexander Keller.
\newblock Instant neural radiance fields.
\newblock In {\em ACM SIGGRAPH 2022 Real-Time Live!} 2022.

\bibitem{nguyen2019hologan}
Thu Nguyen-Phuoc, Chuan Li, Lucas Theis, Christian Richardt, and Yong-Liang
  Yang.
\newblock Hologan: Unsupervised learning of {3D} representations from natural
  images.
\newblock In {\em CVPR}, 2019.

\bibitem{Niemeyer2021Regnerf}
Michael Niemeyer, Jonathan~T. Barron, Ben Mildenhall, Mehdi S.~M. Sajjadi,
  Andreas Geiger, and Noha Radwan.
\newblock {RegNeRF}: Regularizing neural radiance fields for view synthesis
  from sparse inputs.
\newblock In {\em CVPR}, 2022.

\bibitem{Ost_2021_CVPR}
Julian Ost, Fahim Mannan, Nils Thuerey, Julian Knodt, and Felix Heide.
\newblock Neural scene graphs for dynamic scenes.
\newblock In {\em CVPR}, 2021.

\bibitem{ozturkcan2021service}
Selcen Ozturkcan.
\newblock Service innovation: Using augmented reality in the {IKEA Place} app.
\newblock {\em Journal of Information Technology Teaching Cases}, 11(1), 2021.

\bibitem{park2019deepsdf}
Jeong~Joon Park, Peter Florence, Julian Straub, Richard Newcombe, and Steven
  Lovegrove.
\newblock {DeepSDF}: Learning continuous signed distance functions for shape
  representation.
\newblock In {\em CVPR}, 2019.

\bibitem{pathak2016context}
Deepak Pathak, Philipp Krahenbuhl, Jeff Donahue, Trevor Darrell, and Alexei~A
  Efros.
\newblock Context encoders: Feature learning by inpainting.
\newblock In {\em CVPR}, 2016.

\bibitem{philip2018plane}
Julien Philip and George Drettakis.
\newblock Plane-based multi-view inpainting for image-based rendering in large
  scenes.
\newblock In {\em Proceedings of the ACM SIGGRAPH Symposium on Interactive 3D
  Graphics and Games}, pages 1--11, 2018.

\bibitem{rematas2022urban}
Konstantinos Rematas, Andrew Liu, Pratul~P Srinivasan, Jonathan~T Barron,
  Andrea Tagliasacchi, Thomas Funkhouser, and Vittorio Ferrari.
\newblock Urban radiance fields.
\newblock In {\em CVPR}, pages 12932--12942, 2022.

\bibitem{rockwell2021pixelsynth}
Chris Rockwell, David~F Fouhey, and Justin Johnson.
\newblock {PixelSynth}: Generating a {3D}-consistent experience from a single
  image.
\newblock In {\em ICCV}, 2021.

\bibitem{roessle2022dense}
Barbara Roessle, Jonathan~T Barron, Ben Mildenhall, Pratul~P Srinivasan, and
  Matthias Nie{\ss}ner.
\newblock Dense depth priors for neural radiance fields from sparse input
  views.
\newblock In {\em CVPR}, 2022.

\bibitem{runz2020frodo}
Martin Runz, Kejie Li, Meng Tang, Lingni Ma, Chen Kong, Tanner Schmidt, Ian
  Reid, Lourdes Agapito, Julian Straub, Steven Lovegrove, et~al.
\newblock {FroDO}: From detections to {3D} objects.
\newblock In {\em CVPR}, 2020.

\bibitem{saito2019pifu}
Shunsuke Saito, Zeng Huang, Ryota Natsume, Shigeo Morishima, Angjoo Kanazawa,
  and Hao Li.
\newblock {PIFu}: Pixel-aligned implicit function for high-resolution clothed
  human digitization.
\newblock In {\em ICCV}, 2019.

\bibitem{saito2020pifuhd}
Shunsuke Saito, Tomas Simon, Jason Saragih, and Hanbyul Joo.
\newblock {PIFuHD}: Multi-level pixel-aligned implicit function for
  high-resolution {3D} human digitization.
\newblock In {\em CVPR}, 2020.

\bibitem{yu2021plenoxels}
{Sara Fridovich-Keil and Alex Yu}, Matthew Tancik, Qinhong Chen, Benjamin
  Recht, and Angjoo Kanazawa.
\newblock Plenoxels: Radiance fields without neural networks.
\newblock In {\em CVPR}, 2021.

\bibitem{sayed2022simplerecon}
Mohamed Sayed, John Gibson, Jamie Watson, Victor Prisacariu, Michael Firman,
  and Cl{\'e}ment Godard.
\newblock {SimpleRecon}: {3D} reconstruction without {3D} convolutions.
\newblock In {\em ECCV}, 2022.

\bibitem{schonberger2016structure}
Johannes~L Schonberger and Jan-Michael Frahm.
\newblock Structure-from-motion revisited.
\newblock In {\em CVPR}, 2016.

\bibitem{schoenberger2016mvs}
Johannes~Lutz Sch\"{o}nberger, Enliang Zheng, Marc Pollefeys, and Jan-Michael
  Frahm.
\newblock Pixelwise view selection for unstructured multi-view stereo.
\newblock In {\em ECCV}, 2016.

\bibitem{schwarz2020graf}
Katja Schwarz, Yiyi Liao, Michael Niemeyer, and Andreas Geiger.
\newblock Graf: Generative radiance fields for {3D-aware} image synthesis.
\newblock {\em NeurIPS}, 2020.

\bibitem{shih20203d}
Meng-Li Shih, Shih-Yang Su, Johannes Kopf, and Jia-Bin Huang.
\newblock {3D} photography using context-aware layered depth inpainting.
\newblock In {\em CVPR}, 2020.

\bibitem{sucar2021iMap}
Edgar Sucar, Shikun Liu, Joseph Ortiz, and Andrew Davison.
\newblock {iMAP}: Implicit mapping and positioning in real-time.
\newblock In {\em ICCV}, 2021.

\bibitem{sulaiman2020matterport}
Mohamad~Zaidi Sulaiman, Mohd Nasiruddin~Abdul Aziz, Mohd Haidar~Abu Bakar,
  Nur~Akma Halili, and Muhammad~Asri Azuddin.
\newblock Matterport: virtual tour as a new marketing approach in real estate
  business during pandemic {COVID-19}.
\newblock In {\em International Conference of Innovation in Media and Visual
  Design (IMDES)}, 2020.

\bibitem{suvorov2022lama}
Roman Suvorov, Elizaveta Logacheva, Anton Mashikhin, Anastasia Remizova,
  Arsenii Ashukha, Aleksei Silvestrov, Naejin Kong, Harshith Goka, Kiwoong
  Park, and Victor Lempitsky.
\newblock Resolution-robust large mask inpainting with fourier convolutions.
\newblock In {\em WACV}, 2022.

\bibitem{thonat2016multi}
Theo Thonat, Eli Shechtman, Sylvain Paris, and George Drettakis.
\newblock Multi-view inpainting for image-based scene editing and rendering.
\newblock In {\em 3DV}, 2016.

\bibitem{tremblay2022rtmv}
Jonathan Tremblay, Moustafa Meshry, Alex Evans, Jan Kautz, Alexander Keller,
  Sameh Khamis, Charles Loop, Nathan Morrical, Koki Nagano, Towaki Takikawa,
  and Stan Birchfield.
\newblock {RTMV}: A ray-traced multi-view synthetic dataset for novel view
  synthesis.
\newblock {\em ECCV Workshop (Learn3DG ECCVW)}, 2022.

\bibitem{wang20223d}
Qianqian Wang, Zhengqi Li, David Salesin, Noah Snavely, Brian Curless, and
  Janne Kontkanen.
\newblock {3D} moments from near-duplicate photos.
\newblock In {\em CVPR}, 2022.

\bibitem{wang2004image}
Zhou Wang, Alan~C Bovik, Hamid~R Sheikh, and Eero~P Simoncelli.
\newblock Image quality assessment: from error visibility to structural
  similarity.
\newblock {\em IEEE transactions on image processing}, 13(4), 2004.

\bibitem{wu2022object}
Qianyi Wu, Xian Liu, Yuedong Chen, Kejie Li, Chuanxia Zheng, Jianfei Cai, and
  Jianmin Zheng.
\newblock Object-compositional neural implicit surfaces.
\newblock In {\em ECCV}, 2022.

\bibitem{xu2019VideoInpainting}
Rui Xu, Xiaoxiao Li, Bolei Zhou, and Chen~Change Loy.
\newblock Deep flow-guided video inpainting.
\newblock In {\em CVPR}, 2019.

\bibitem{xu2019view}
Xiaogang Xu, Ying-Cong Chen, and Jiaya Jia.
\newblock View independent generative adversarial network for novel view
  synthesis.
\newblock In {\em ICCV}, 2019.

\bibitem{yang2021objectnerf}
Bangbang Yang, Yinda Zhang, Yinghao Xu, Yijin Li, Han Zhou, Hujun Bao, Guofeng
  Zhang, and Zhaopeng Cui.
\newblock Learning object-compositional neural radiance field for editable
  scene rendering.
\newblock In {\em ICCV}, October 2021.

\bibitem{yu2021plenoctrees}
Alex Yu, Ruilong Li, Matthew Tancik, Hao Li, Ren Ng, and Angjoo Kanazawa.
\newblock {PlenOctrees} for real-time rendering of neural radiance fields.
\newblock In {\em ICCV}, 2021.

\bibitem{yu2018generative}
Jiahui Yu, Zhe Lin, Jimei Yang, Xiaohui Shen, Xin Lu, and Thomas~S Huang.
\newblock Generative image inpainting with contextual attention.
\newblock In {\em CVPR}, 2018.

\bibitem{yuan2022nerf}
Yu-Jie Yuan, Yang-Tian Sun, Yu-Kun Lai, Yuewen Ma, Rongfei Jia, and Lin Gao.
\newblock {NeRF}-editing: geometry editing of neural radiance fields.
\newblock In {\em CVPR}, 2022.

\bibitem{zeng2020learning}
Yanhong Zeng, Jianlong Fu, and Hongyang Chao.
\newblock Learning joint spatial-temporal transformations for video inpainting.
\newblock In {\em ECCV}, 2020.

\bibitem{zeng2019learning}
Yanhong Zeng, Jianlong Fu, Hongyang Chao, and Baining Guo.
\newblock Learning pyramid-context encoder network for high-quality image
  inpainting.
\newblock In {\em CVPR}, 2019.

\bibitem{zhang2018perceptual}
Richard Zhang, Phillip Isola, Alexei~A Efros, Eli Shechtman, and Oliver Wang.
\newblock The unreasonable effectiveness of deep features as a perceptual
  metric.
\newblock In {\em CVPR}, 2018.

\bibitem{Zhi2021Inplace}
Shuaifeng Zhi, Tristan Laidlow, Stefan Leutenegger, and Andrew Davison.
\newblock In-place scene labelling and understanding with implicit scene
  representation.
\newblock In {\em ICCV}, 2021.

\end{thebibliography}
